\PassOptionsToPackage{dvipsnames}{xcolor}
\documentclass[11pt]{article}
\usepackage{emnlp2021}
\usepackage{times}
\usepackage{latexsym}

\usepackage{microtype}

\usepackage{array}
\usepackage{amsmath}
\usepackage{arydshln}
\usepackage{multirow}
\usepackage{graphics}
\usepackage{comment}
\usepackage{url}
\usepackage{textcomp}
\usepackage{mathtools}
\usepackage{xcolor}
\usepackage{makecell}
\usepackage{tikz}

\usepackage{amsmath,amsfonts,bm}

\def\eqref#1{equation~\ref{#1}}

\def\1{\bm{1}}

\def\eps{{\epsilon}}

\DeclareMathAlphabet{\mathsfit}{\encodingdefault}{\sfdefault}{m}{sl}
\SetMathAlphabet{\mathsfit}{bold}{\encodingdefault}{\sfdefault}{bx}{n}

\DeclareMathOperator{\emb}{x}
\DeclareMathOperator{\score}{\phi}

\usepackage{tikz}
\usetikzlibrary{calc}

\title{Towards Zero-shot Commonsense Reasoning \\with Self-supervised Refinement of Language Models}

\author{Tassilo Klein \\
  SAP AI Research \\
  \texttt{tassilo.klein@sap.com} \\\And
  Moin Nabi \\
    SAP AI Research \\
  \texttt{m.nabi@sap.com} \\}

\date{}

\begin{document}
\maketitle
\begin{abstract}
Can we get existing language models and refine them for zero-shot commonsense reasoning?
This paper presents an initial study exploring the feasibility of zero-shot commonsense reasoning for the Winograd Schema Challenge by formulating the task as self-supervised refinement of a pre-trained language model. In contrast to previous studies that rely on fine-tuning annotated datasets, we seek to boost conceptualization via loss landscape refinement. To this end, we propose a novel self-supervised learning approach that refines the language model utilizing a set of linguistic perturbations of similar concept relationships. Empirical analysis of our conceptually simple framework demonstrates the viability of zero-shot commonsense reasoning on multiple benchmarks.\footnote{The source code can be found at: \url{https://github.com/SAP-samples/emnlp2021-contrastive-refinement/}}

\end{abstract}

\section{Introduction}
Natural language processing has recently experienced unprecedented progress, boosting the performance of many applications to new levels. However, this gain in performance does not equally transfer to applications requiring commonsense reasoning capabilities, which has largely remained an unsolved problem~\cite{marcus2020next,kocijan2020review}.
In order to assess the commonsense reasoning capabilities of automatic systems, several tasks have been devised. Among them is the popular Winograd Schema Challenge (WSC), which frames commonsense reasoning as a pronoun resolution task~\cite{levesque2012winograd}. Although appearing evident and natural to the human mind, modern machine learning methods still struggle to solve this challenge. \\
Lately, the research community has experienced an abundance of methods proposing utilization of language models (LM) to tackle commonsense reasoning in a two-stage learning pipeline. Starting from an initial self-supervised learned model, commonsense enhanced LMs are obtained in a subsequent fine-tuning (ft) phase. Fine-tuning enforces the LM to solve the downstream WSC task as a plain co-reference resolution task.
However, such supervised approaches are prone to leverage statistical data artifacts for reasoning, giving rise to the ``Clever Hans'' effect~\cite{lapuschkin-ncomm19}. As such, instead of truly featuring reasoning capabilities, approaches become very good in faking. On the other hand, the lack of commonsense reasoning capabilities of LMs can be partially attributed to the training corpora itself, as the commonsense knowledge is often not incorporated into the training text due to the assumed triviality~\cite{trichelair2018evaluation,DBLP:journals/corr/abs-1810-00521,trichelair-etal-2019-reasonable,emami2019knowref, kavumba2019choosing, liu2020precise,cui2020does}.
\begin{figure}
    \centering
    \includegraphics[width=0.49\textwidth]{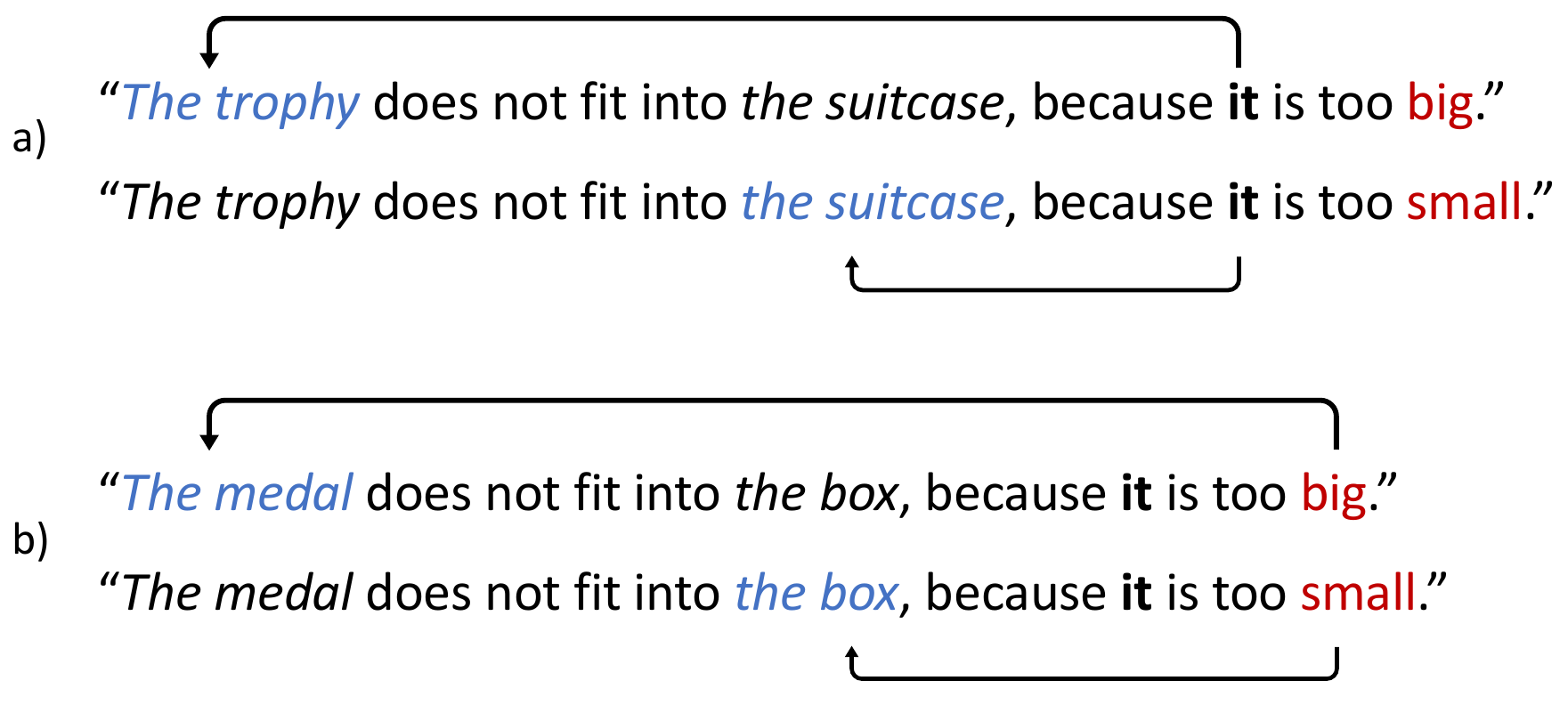}
    \caption{WSC sample: a) original sentence, b) perturbation (noun synonym). Task: resolve \textcolor{black}{\textbf{pronoun}} with a \textcolor{NavyBlue}{candidate}. The \textcolor{Maroon}{trigger-word} induces an answer flip.}
\label{fig:sentences}
\end{figure}
We hypothesize that the current self-supervised tasks used in the pre-training phase are insufficient to enforce the model to generalize commonsense concepts~\cite{kejriwal2020finetuned}. This shortcoming is easily unveiled by the susceptibility of LM to semantic variations. In this regard, it has been shown that LMs are sensitive to linguistic perturbations~\cite{abdou-etal-2020-sensitivity}.
A case in point is the WSC example in Fig.~\ref{fig:sentences}. It shows a pair of sentences subject to semantic variations establishing the same relationship between entities. This can be defined as the joint concept triplet consisting of two nouns and a verb that determines the relationship between the nouns, e.g., {\tt{(container, item, fit)}}. Inappropriate semantic sensitivity to semantic variants leads to inadequate ``conceptualization'' and misconstruction of such triplets. To address this, we propose \emph{self-supervised refinement}. It seeks to achieve generalization through a task agnostic objective.

To this end, we tackle the problem of commonsense reasoning from a zero-shot learning perspective. Leveraging zero-shot models to gauge the intrinsic incorporation of commonsense knowledge suggests being the more valid approach than fine-tuned models. That can be attributed to the exploitation of implicit biases less likely to occur in this setup. Hence, the associated benchmarks constitute a more realistic and reliable benchmark~\cite{elazar2021square}. Other zero-shot methods for commonsense reasoning either use large supervised datasets {\tt{Winogrande}}~\cite{WinoGrande}) or very large LMs such as {\tt{GPT-3}}~\cite{GPT3}.
In contrast, the proposed method takes a pre-trained LM as input, which undergoes a refinement step. During refinement, the LM is exposed to semantic variations, aiming at improved concept generalization by making the model more robust w.r.t. perturbations.
Motivated by the recent advancements in contrastive representation learning~\cite{chen2020simple,He_2020_CVPR,jean-bastien2020bootstrap,klein2020contrastive}, we propose refining the LM in a self-supervised contrastive fashion. This entails refinement without the use of any labels and hence with \emph{no gradient update} on the downstream datasets. Consequently, the supervision level is identical to the test time of the Winograd schema challenge.

Our contributions are two-fold: \textbf{(i)} we introduce the task of \emph{zero-shot commonsense reasoning} for WSC by reformulating the task as performing self-supervised refinement on a pre-trained language model \textbf{(ii)} We propose a self-supervised refinement framework which leverages semantic perturbations to facilitate zero-shot commonsense reasoning.

\section{Method}
 \begin{figure*}[t!]
    \centering
    \includegraphics[width=1\textwidth]{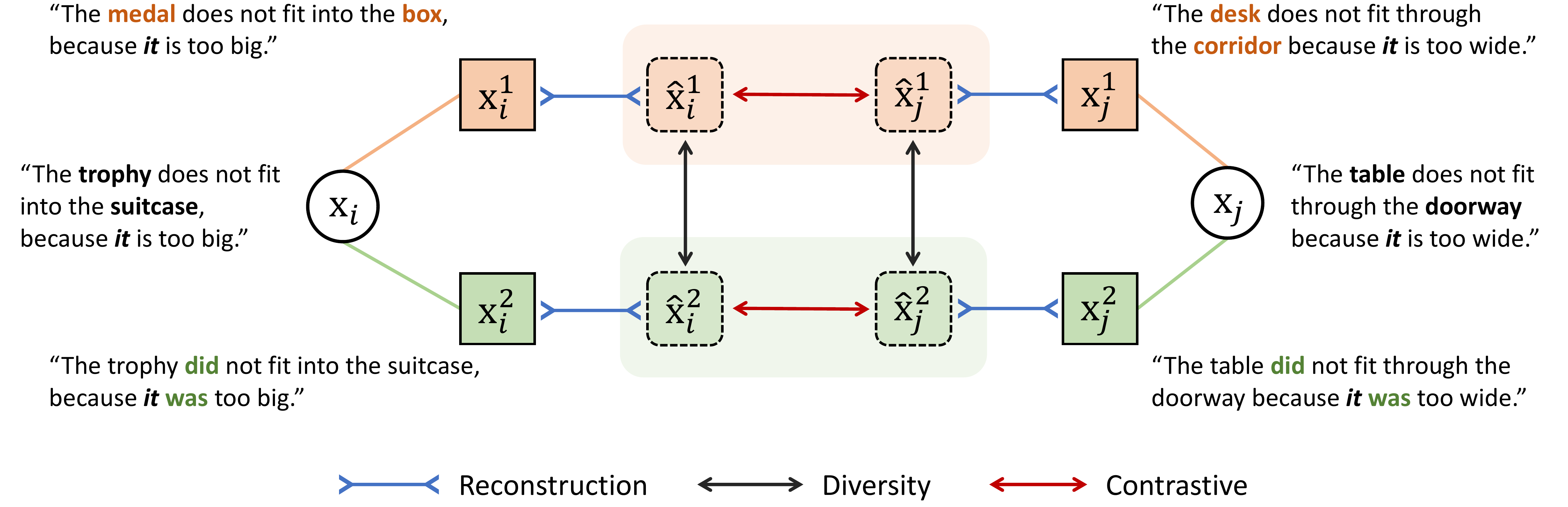}
    \caption{Schematic illustration of the proposed approach. Two examples $\emb_{i}$ and $\emb_{j}$ from the WSC dataset, both demonstrating the concept triplet {\tt{(container, item, fit)}} and their generated embeddings (dashed outline) for two perturbation types: \textbf{top:} \textbf{\textcolor{RawSienna}{\tt[SYNOYM]}} and \textbf{bottom:} \textbf{\textcolor{ForestGreen}{\tt{[TENSE]}}}. Loss terms defined as \emph{attraction} \textbf{($\longleftrightarrow$)} and \emph{repulsion}
\textbf{(\begin{tikzpicture}[transform canvas={yshift=59.5pt}]
    \draw [shorten >=0.3cm, stealth reversed-stealth reversed](0,-2) -- (1,-2);\end{tikzpicture}\quad\quad)}
    between embeddings of unperturbed and corresponding generated perturbation, each shown in a different color: \textcolor{NavyBlue}{\textbf{Reconstruction}} loss, \textcolor{Maroon}{\textbf{Contrastive}} loss and \textcolor{black}{\textbf{Diversity}} loss (best shown in color). }
\label{fig:method}
\end{figure*}
\noindent\textbf{Preliminaries: }
Transformer-based LMs are based on an encoder-decoder architecture, consisting of a set of encoding layers that process the input iteratively. Prior to entering the Transformer stack, the input is pre-processed by a tokenizer that turns the input sentence into a sequence of tokens.
Besides tokens arising from the input sentence, there are also auxiliary tokens such as {\tt [CLS],{\tt [SEP]}}. In BERT and RoBERTa,  these tokens delimit the input from padding for fixed-length sequence processing. Furthermore, there are special tokens that are tailored to frame specific tasks. For example,  {\tt [MASK]} is used to mask out words for learning the masked language model. Instantiation of language models on the tokenized sequence yields a sequence of embedding vectors.
To avoid clutter in the notation and subsuming the fact that only fixed-length sequences are encoded, for the following $\emb \in \mathbb{T}$ will refer to the tensor obtained by stacking the sequence of token embeddings. \subsection{Perturbation Generation Framework} Starting from a pre-trained LM ({\tt{init-LM}}), we conduct a refinement step exposing the model to semantic variations of Winograd schemas. Given a sentence $x$ and a specific semantic {\tt{[perturbation token]}}, the LM is trained to generate the embedding $\hat{x}$ of the provided perturbation type.
We enforce the generator to estimate the embedding obtained by the LM on the sentence with the actual semantic perturbation as the target.
Intuitively speaking, an LM that generates perturbed representations from an unperturbed input is equipped with a generalized view of commonsense concepts. This builds upon the idea that the injection of noise to the input can flatten the loss landscape to promote generalization~\cite{NEURIPS2019_0defd533,Moosavi-Dezfooli_2019_CVPR}.

To this end, we extend the set of auxiliary tokens with some new tokens referred to as ``perturbation tokens''.  In the course of training, the perturbation tokens are prepended to the input sentence directly after the {\tt[CLS]} token.
For the following, we let $\mathcal{P}$ denote the set of semantic perturbations. Besides perturbations, $\mathcal{P}$ also includes the identity transformation {\tt [IDENTICAL]}, which implies no semantic change. Figure~\ref{fig:sentences} shows an example of a perturbation induced by the perturbation token {\tt [SYNONYM]}, which entails replacing nouns of the input sentence with synonyms. Following the example from the figure, the LM seeks to map the representation of the (tokenized) sentence \textbf{(a)} in conjunction with {\tt [SYNONYM]} to the representation of \textbf{(b)}.
To enforce consistency across commonsense concepts and semantic perturbations, we embed learning in a contrastive setting.

\subsection{Self-supervised Refinement}
The method's core idea is to construct an abstract, generic view of a commonsense concept by exploiting slightly different examples of the same concept (i.e., perturbations). This is achieved by joint optimization of a LM w.r.t. three different loss terms (Reconstruction, Contrastive and Diversity):
\begin{equation}
\label{eq:NZG}
\begin{aligned}
\min_{\theta_1,\theta_2} \mathcal{L}_{R}(f_{\theta_1}) + \mathcal{L}_{C}(f_{\theta_1}) + \mathcal{L}_{D}(f_{\theta_1},q_{\theta_2})
\end{aligned}
\end{equation}
Here $f$ denotes the LM, e.g., BERT or RoBERTa parameterized by $\theta_1$, and $q: \mathbb{T} \to \mathcal{P}$ denotes a representation discriminator (MLP) parameterized by $\theta_2$.
The functionality of the individual loss terms of Eq.~\ref{eq:NZG} will be explained in the following subsections. Additionally, Fig.~\ref{fig:method} shows a schematic illustration of the proposed approach and each loss term.
\\
Optimization of Eq.~\ref{eq:NZG} entails computation of similarities between embeddings, employing a metric $\score(\emb, \hat{\emb}):  \mathbb{T} \times \mathbb{T} \to \mathbb{R}$. Here, we employ a variant of the BERTscore~\cite{bert-score} as a similarity metric.  BERTscore computes sentence similarities by matching tokens based on their cosine similarity. Subsequently, the scores for the entire sequence are aggregated. Unlike the original BERTscore, we restrict token matching to each token's vicinity to accommodate that perturbations typically induce changes only in a small neighborhood. To this end, we restrict token matching by applying a sliding window mechanism centered on each token.
\\
\subsubsection{Reconstruction loss}
The reconstruction loss's objective is to regress embeddings by minimizing the distance between the ground-truth and the approximated ``perturbation'' embedding:
\vspace{-2mm}
\begin{equation}
   \mathcal{L}_{R}=-\alpha\sum_{i}^N \sum_{k \in \mathcal{P}}  \score(\emb_{i}^{[k]}, \hat{\emb}_{i}^{[k]})
   \label{eq:reconstruction_loss}
\end{equation}

\subsubsection{Contrastive loss}

The objective of the contrastive loss is to preserve the  ``semantic expressivity of individual samples and prevent the collapse to a singular perturbation representation. This is achieved by pushing apart the embeddings for \emph{different samples} of the \emph{same perturbation} type. \vspace{-2mm}
\begin{equation}
   \mathcal{L}_{C}=\beta\sum_{i,j: i\neq j}^N \sum_{k \in \mathcal{P}}  \score(\hat{\emb}_{i}^{[k]}, \hat{\emb}_{j}^{[k]})
   \label{eq:constrastive_loss}
\end{equation}

\subsubsection{Diversity loss}
The diversity loss term aims to guarantee the discriminativeness of the perturbation embeddings arising from the same sample. As such, it imposes the semantic perturbations for the same sample to be diverse, preventing the collapse of different perturbations to a single embedding. Maximizing diversity entails minimization of cross-entropy w.r.t. perturbations:
\vspace{-2mm}
\begin{equation}
     \mathcal{L}_{D}=-\gamma\sum_{i}^N\sum_{k \in \mathcal{P}}  \log \frac{q(k|\hat{\emb}_{i}^{[k]})}{\sum_{\forall t\in \mathcal{P}:t\neq k}q(t|\hat{\emb}_{i}^{[k]})},
     \label{eq:diversity_loss}
\end{equation}

Here $q(.|.): \mathbb{T} \to \mathbb{R}$ denotes the likelihood of a classifier w.r.t. embeddings. $N$ denotes the number of data samples, and $\alpha,\beta,\gamma \in \mathbb{R}$ denote the hyperparameters, balancing the terms in the loss function.

\subsubsection{Zero-shot Pronoun Disambiguation}
For resolving the WSC we leverage the Transformer Masked Token Prediction following~\cite{kocijan19acl}. This entails replacing the {\tt [MASK]} token with the possible candidates. Given an associated pair of training sentences with $i
\in N$, i.e., $\left(s_i^1, s_i^2\right)$, the difference between the sentence pairs is the trigger word(s). Here $c_1, c_{2}$ denote the answer candidates, yielding probabilities for the candidates:
$p\left(c_1|s_i^1\right)$ and $p\left(c_2|s_i^1\right)$.
The answer prediction corresponds to the candidate with a more significant likelihood. If a candidate consists of several tokens, the probability corresponds to the average of the log probabilities.
\begin{table}[t!]
\begin{center}
\begin{minipage}{2.6in}
\begin{tabular}{l|ll}
\hline
\multicolumn{3}{c}{\textbf{DPR}~\cite{rahman-ng-2012-resolving}} \\
\hline
Method & BERT & RoBERTa\\
\hline
Baseline (\texttt{init-LM})
& 58.50 \% & 70.39 \%\\
\hdashline
Ours (\texttt{Zero-shot})
& \textbf{61.35 \%} & \textbf{76.95 \%}  \\
\hline \hline
\end{tabular}
\vspace{1mm}
\begin{tabular}{l|ll}
\multicolumn{3}{c}{\textbf{GAP}~\cite{webster-etal-2018-mind}} \\
\hline
Method & BERT & RoBERTa\\
\hline
Baseline (\texttt{init-LM})
& 58.70 \% & 58.87 \%\\
\hdashline
Ours (\texttt{Zero-shot})
& \textbf{58.73 \%} & \textbf{59.13 \%}  \\
\hline \hline
\end{tabular}
\vspace{1mm}
\begin{tabular}{l|ll}
\multicolumn{3}{c}{\textbf{KnowRef}~\cite{emami2019knowref}} \\
\hline
Method & BERT & RoBERTa\\
\hline
Baseline (\texttt{init-LM})
& 62.36 \%  & 60.42 \%\\
\hdashline
Ours (\texttt{Zero-shot})
& \textbf{62.44 \%} &  \textbf{63.97 \%}  \\
\hline \hline
\end{tabular}
\vspace{1mm}
\begin{tabular}{l|ll}
\multicolumn{3}{c}{\textbf{PDP-60}~\cite{davis2016human}} \\
\hline
Method & BERT & RoBERTa\\
\hline
Baseline (\texttt{init-LM})
& \textbf{60.00 \%}  & 50.00 \%\\
\hdashline
Ours (\texttt{Zero-shot})
& 58.33 \% &  \textbf{55.00 \%}  \\
\hline \hline
\end{tabular}
\vspace{1mm}
\begin{tabular}{l|ll}
\multicolumn{3}{c}{\textbf{WSC-273}~\cite{levesque2012winograd}} \\
\hline
Method & BERT & RoBERTa\\
\hline
Baseline (\texttt{init-LM})
& \textbf{62.64 \%}  & 67.77 \% \\
\hdashline
Ours (\texttt{Zero-shot})
& 61.54 \% &  \textbf{71.79 \%}  \\
\hline \hline
\end{tabular}
\vspace{1mm}
\begin{tabular}{l|ll}
\multicolumn{3}{c}{\textbf{WinoGender}~\cite{rudinger-etal-2018-gender}} \\
\hline
Method & BERT & RoBERTa\\
\hline
Baseline (\texttt{init-LM})
& \textbf{62.50 \%}  & 61.67 \%\\
\hdashline
Ours (\texttt{Zero-shot})
& 62.08 \% &  \textbf{69.17 \%}   \\
\hline \hline
\end{tabular}
\vspace{1mm}
\begin{tabular}{l|ll}
\multicolumn{3}{c}{\textbf{WinoGrande}~\cite{WinoGrande}} \\
\hline
Method & BERT & RoBERTa\\
\hline
Baseline (\texttt{init-LM})
& 51.70 \% & 53.75 \% \\
\hdashline
Ours (\texttt{Zero-shot})
& \textbf{52.33 \%} &  \textbf{55.01 \%}  \\
\hline \hline
\end{tabular}
\vspace{1mm}
\begin{tabular}{l|ll}
\multicolumn{3}{c}{\textbf{WinoBias Anti}~\cite{zhao-etal-2018-gender}} \\
\hline
Method & BERT & RoBERTa\\
\hline
Baseline (\texttt{init-LM})
& \textbf{56.82} \% & 55.93 \%\\
\hdashline
Ours (\texttt{Zero-shot})
& \textbf{56.82 \%} &  \textbf{60.61 \%}  \\
\hline \hline
\end{tabular}
\vspace{1mm}
\begin{tabular}{l|ll}
\multicolumn{3}{c}{\textbf{WinoBias Pro}~\cite{zhao-etal-2018-gender}} \\
\hline
Method & BERT & RoBERTa\\
\hline
Baseline (\texttt{init-LM})
& 68.43 \% & 68.43 \% \\
\hdashline
Ours (\texttt{Zero-shot})
& \textbf{75.12 \%} &  \textbf{75.76 \%}\\
\hline
\end{tabular}
\end{minipage}
\caption{Results for zero-shot commonsense reasoning }
\label{tab:zs-results}
\hfill
\end{center}
\end{table}

\section{Experiments and Results}
\subsection{Setup}
We approach training the language model by first training the LM on perturbations on the enhanced-WSC corpus~\cite{abdou-etal-2020-sensitivity}. It is a perturbation augmented version of the original WSC dataset.  It consists of $285$ sample sentences, with up to $10$ semantic perturbations per sample. We make use of the following 7 perturbations:  tense switch {\tt{[TENSE]}}, number switch {\tt{[NUMBER]}}, gender switch {\tt{[GENDER]}}, voice switch (active to passive or vice versa) {\tt{[VOICE]}}, relative clause insertion (a relative clause is inserted after the first referent){\tt{[RELCLAUSE]}}, adverbial qualification (an adverb is inserted to qualify the main verb of each instance){\tt{[ADVERB]}}, synonym/name substitution {\tt{[SYNONYM]}}.
\\
\subsection{Architecture}
The proposed approach is applicable to any Transformer architecture. Here, we adopted standard LMs such as BERT and RoBERTa for comparability, without aiming to optimize the results for any downstream dataset/benchmark. Specifically, we employ the Hugging Face~\cite{Wolf2019HuggingFacesTS} implementation of BERT \emph{large-uncased} architecture as well as RoBERTA \emph{large}.
The LM is trained for $10$ epochs for BERT and $5$ for RoBERTa, using a batch size of $10$ sentence samples. Each sample was associated with $4$ perturbation, yielding an effective batch size of $40$. For optimization, we used a typical setup of AdamW with $500$ warmup steps, a learning rate of $5.0^{-5}$ with $\eps=1.0^{-8}$ and $\eps=1.0^{-5}$ for BERT and RoBERTa, respectively.
For training BERT, we used $\alpha=130$, $\beta=0.5$, $\gamma=2.5$, for RoBERTa $\alpha=1.25$, $\beta=7.25$, $\gamma=6.255$. For hyperparameter optimization of $\alpha, \beta, \gamma$ we follow a standard greedy heuristic, leveraging a weighted-sum optimization scheme~\cite{JakobBlume2014}. From an initial a candidate solution set, coarse-grid random search is utilized to explore the neighborhood on a fine grid of a randomly selected candidates.

The representation discriminator $q$ is a MLP consisting of two fully connected layers with BatchNorm, parametric ReLU (PReLU) activation function and $20\%$ Dropout.

\subsection{Results}
 Given the BERT and RoBERTa language models for comparison, the baseline constitute the {\tt{initial-LM}} prior to undergoing refinement. We evaluated our method on nine different benchmarks. Results are reported in Tab. \ref{tab:zs-results}.
Accuracy gains are significant and consistent with RoBERTa across all benchmarks. On average, the proposed approach increases the accuracy of $(+0.8\%)$ with BERT and of $(+4.5\%)$ with RoBERTa.  The benchmarks and the results are discussed below:
\\
\noindent\textbf{\emph{DPR}}~\cite{rahman-ng-2012-resolving}: a pronoun disambiguation benchmark resembling WSC-273, yet significantly larger. According to ~\cite{trichelair2018evaluation}, less challenging due to inherent biases. Here the proposed approach outperforms the baseline for both BERT and RoBERTA by a margin of $(+2.85\%)$ and $(+6.56\%)$, respectively.
\\
\noindent\textbf{\emph{GAP}}~\cite{webster-etal-2018-mind}: a gender-balanced co-reference corpus. The proposed approach outperforms the baseline on BERT and RoBERTA with $(+0.08\%)$ and $(+0.26\%)$.
\\
\noindent\textbf{\emph{KnowRef}}~\cite{emami2019knowref}: a co-reference corpus addressing gender and number bias. The proposed approach outperforms the baseline on BERT and RoBERTA with $(+0.08\%)$ and $(+3.55\%)$.
\\
\noindent\textbf{\emph{PDP-60}}~\cite{davis2016human}: pronoun disambiguation dataset. Our method outperforms the baseline with RoBERTa with (+5.0\%), while on BERT showing a drop of (-1.67\%).
\\
\noindent\textbf{\emph{WSC-273}}~\cite{levesque2012winograd}: a pronoun disambiguation benchmark, known to be more challenging than PDP-60. Our method outperforms the baseline with RoBERTa with $(+4.0\%)$, with a drop of $(-1.1\%)$ with BERT.
\\
\noindent\textbf{\emph{WinoGender}}~\cite{rudinger-etal-2018-gender}: a gender-balanced co-reference corpus. The proposed approach outperforms the baseline on RoBERTA by $(+7.6\%)$, observing a drop on BERT $(-0.42\%)$.
\\
\noindent\textbf{\emph{WinoGrande (W.G.)}}~\cite{WinoGrande}: the largest dataset for Winograd commonsense reasoning. Our method outperforms the baseline with BERT by $(+0.63)$ and with RoBERTa by $(+1.26\%)$.
\\
\noindent\textbf{\emph{WinoBias}}~\cite{rudinger-etal-2018-gender}: a gender-balanced co-reference corpus consisting of two types. Type-1 requiring world knowledge, Type-2 requiring syntactic understanding. While on par for the first type in combination with BERT and a margin of $(+6.69\%)$, the proposed approach outperforms the baseline with RoBERTa with $(+4.68)$ and $(+7.33)$.
\\
\vspace{-2mm}
\subsubsection{Ablation Study}
To assess each loss term's contribution, we evaluated each component's performance by removing them individually from the loss. It should be noted that $\mathcal{L}_{C}$ of Eq.~\ref{eq:constrastive_loss} and $\mathcal{L}_{D}$ of Eq.~\ref{eq:diversity_loss} both interact in a competitive fashion. Hence, only the equilibrium of these terms yields an optimal solution. Changes - such as eliminating a term - have detrimental effects, as they prevent achieving such an equilibrium, resulting in a significant drop in performance. See Tab. \ref{tab:ablation-study} for the ablation study on two benchmarks. Best performance is achieved in the presence of all loss terms.

\begin{table}
\centering
\begin{tabular}{l|ll}
\hline
\textbf{Method} & \textbf{DPR} & \textbf{W.G.}\\
\hline
\hline
\centering
RoBERTa~\cite{liu2019roberta} &70.39 & 53.75\\
\hdashline
Ours (\texttt{$\mathcal{L}_{C}$+$\mathcal{L}_{D}$}) & 73.76 & 53.28\\
Ours (\texttt{$\mathcal{L}_{R}$+$\mathcal{L}_{D}$}) & 65.60 & 53.59\\
Ours (\texttt{$\mathcal{L}_{R}$+$\mathcal{L}_{C}$}) & 65.07 & 52.01\\
\textbf{Ours} (\texttt{$\mathcal{L}_{R}$+$\mathcal{L}_{C}$+$\mathcal{L}_{D}$}) & \textbf{76.95} & \textbf{55.01} \\
\hline
\end{tabular}
\caption{Ablation study, performance in accuracy (\%)}
\label{tab:ablation-study}
\end{table}

\section{Discussion and Conclusion}
We introduced a method for self-supervised refinement of LMs. Its conceptual simplicity facilitates generic integration into frameworks tackling commonsense reasoning.
A first empirical analysis on multiple benchmarks indicates that the proposed approach consistently outperforming the baselines in terms of standard pre-trained LMs, confirming the fundamental viability. We believe that the performance gain will be more pronounced when leveraging larger perturbation datasets for LM refinement. Hence, future work will focus on the generation of perturbations. This could specifically entail the consideration of sample-specific perturbations.

\bibliographystyle{acl_natbib}
\bibliography{anthology,emnlp2021}

\end{document}